\documentclass[10pt, letter, onecolumn]{arxiv}

\usepackage{kantlipsum, lipsum}
\usepackage{dm-colors}
\usepackage{amsmath}
\usepackage{pstricks, pst-node}
\usepackage{verbatim}
\usepackage{multirow}
\usepackage{array}
\usepackage{longtable}
\usepackage{pdflscape}
\usepackage{scalerel}
\usepackage{booktabs}
\usepackage{enumitem}
\usepackage{xspace}
\usepackage{bm}
\usepackage{bbm}
\usepackage{mathtools}
\usepackage{soul}
\usepackage{epsfig}
\usepackage{graphicx}
\usepackage{amssymb}
\usepackage{colortbl}
\usepackage{csquotes}
\usepackage{setspace}
\usepackage{colortbl}
\usepackage{bbding}
\usepackage{threeparttable}
\usepackage{tabularx,ragged2e}
\usepackage{placeins}
\usepackage{bbding}
\usepackage{wrapfig}

\definecolor{darkpastelgreen}{rgb}{0.13, 0.55, 0.13}
\definecolor{darkpastelred}{rgb}{0.55, 0.13, 0.13}
\definecolor{mygray}{rgb}{1, 1, 1}
\usepackage{xcolor} 

\usepackage[hang,flushmargin]{footmisc}

\usepackage{nameref}
\usepackage{varioref}
\usepackage{amssymb}
\usepackage{pifont}
\usepackage{rotating}
\usepackage{tabularx}
\usepackage[pagebackref=false,breaklinks=false,%
            colorlinks=true,bookmarks=true,citecolor=ourdarkblue,%
            urlcolor=ourdarkblue,linkcolor=ourdarkblue]{hyperref}
\usepackage[noabbrev,capitalize]{cleveref}
\usepackage{etoc}
\usepackage{lineno}
\usepackage{algorithm}
\usepackage{algpseudocode}

\graphicspath{{figures/}}

\title{\Large{ChestX-Reasoner: Advancing Radiology Foundation Models with Reasoning through Step-by-Step Verification}}

\author[1,2, $\ast$]{Ziqing Fan}
\author[2, $\ast$]{Cheng Liang}
\author[1,2]{Chaoyi Wu} 
\author[1,2]{\\ \vspace{0.1cm} Ya Zhang}
\author[1,2]{Yanfeng Wang} 
\author[1,2]{Weidi Xie}

\affil[1]{\normalsize Shanghai Jiao Tong University, Shanghai, China \authorcr \vspace{0.1cm}}

\affil[2]{\normalsize Shanghai Artificial Intelligence Laboratory, Shanghai, China  \authorcr \vspace{0.1cm}
}
\affil[$\ast$]{\normalsize Equal contributions\hspace{1cm}}

\begin{document}

\begin{abstract}
Recent advances in reasoning-enhanced large language models (LLMs) and multimodal LLMs (MLLMs) have significantly improved performance in complex tasks, yet medical AI models often overlook the structured reasoning processes inherent in clinical practice. 
In this work, we present \textbf{ChestX-Reasoner}, 
a radiology diagnosis MLLM designed to leverage process supervision mined directly from clinical reports, reflecting the step-by-step reasoning followed by radiologists. We construct a large dataset by extracting and refining reasoning chains from routine radiology reports. 
Our two-stage training framework combines supervised fine-tuning and reinforcement learning guided by process rewards to better align model reasoning with clinical standards. 
We introduce \textbf{RadRBench-CXR}, a comprehensive benchmark featuring 59K visual question answering samples with 301K clinically validated reasoning steps, and propose \textbf{RadRScore}, a metric evaluating reasoning factuality, completeness, and effectiveness. 
ChestX-Reasoner outperforms existing medical and general-domain MLLMs in both diagnostic accuracy and reasoning ability,
achieving 16\%, 8.5\%, and 18\% improvements in reasoning ability compared to the best medical MLLM, the best general MLLM, and its base model, respectively, as well as 3.3\%, 24\%, and 27\% improvements in outcome accuracy. All resources are open-sourced to facilitate further research in medical reasoning MLLMs.
\end{abstract}

\maketitle


\renewcommand\labelitemi{{\boldmath$\circ$}}

\begin{figure*}[!pht]
    \centering
    \includegraphics[width=0.98\linewidth]{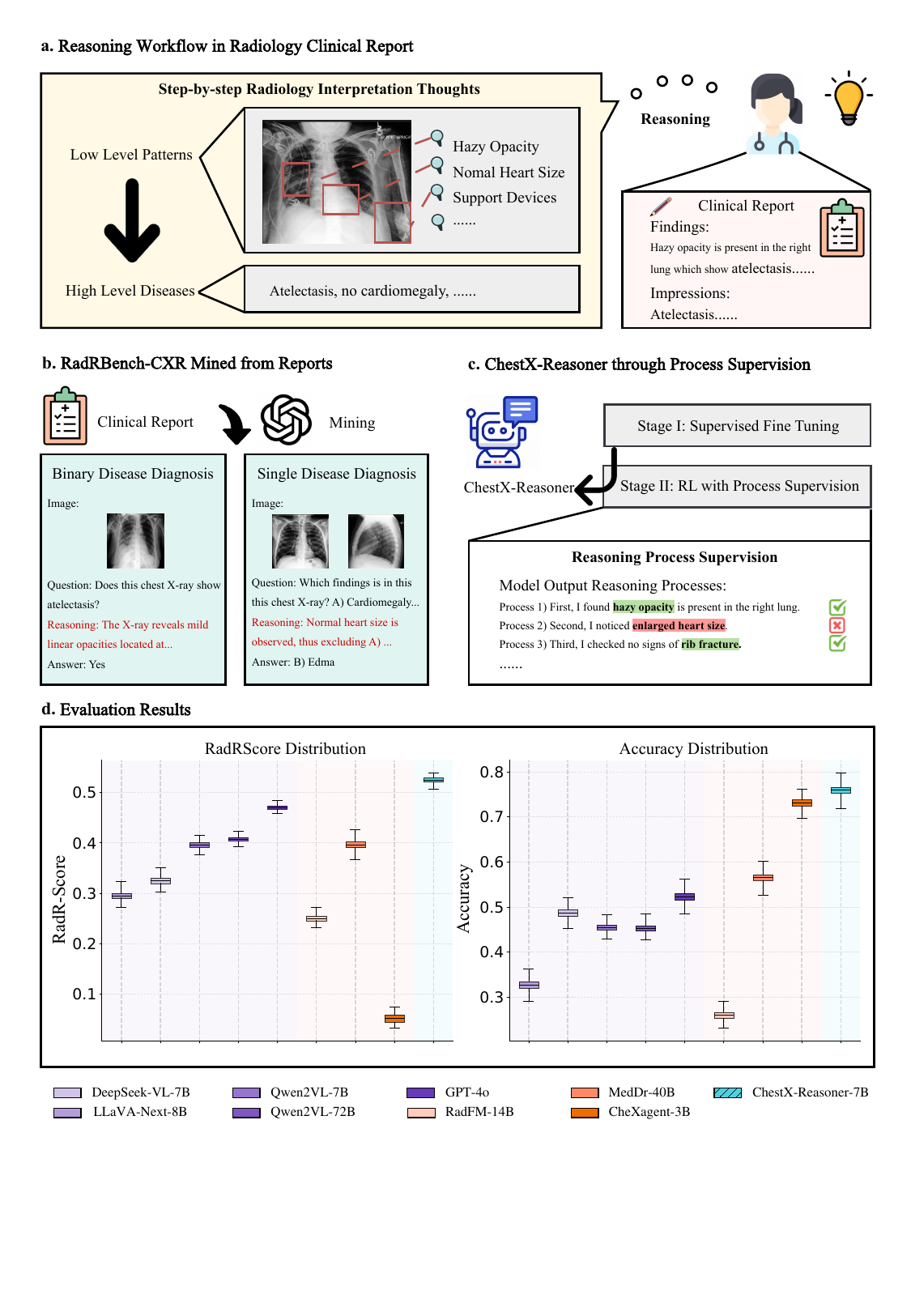}
    \vspace{2pt}
    \caption{\textbf{Mining Reasoning from Clinical Reports for Enhancing MLLMs.} a. The reasoning workflow in radiologists' clinical report writing from low-level patterns of clinical observations to high-level disease identification. b. Mined reasoning samples from clinical reports in conjunction with GPT-4o. c. A two-stage training pipeline that incorporates process supervision to develop our reasoning-enhanced MLLM ChestX-Reasoner. d. An overview of evaluation results on RadRBench-CXR in terms of both reasoning ability and outcome accuracy, compared to extensive baseline models.}
    \label{fig:intro}
    \vspace{-15pt}
\end{figure*}


\section{Introduction}
In the recent literature, the reasoning-enhanced large language models (LLMs), 
such as OpenAI-o3~\cite{openaio3} and DeepSeek-R1~\cite{deepseekr1}, 
have achieved tremendous success in tasks like mathematics and programming~\cite{deepseekmath1,deepseekmath2,gptmath}. Similarly, advancements in multimodal large language models~(MLLMs) have been demonstrated in reasoning tasks~\cite{vlmr2, vlmr1}. Inspired by the success of chain-of-thoughts (CoTs), these models aim to generate intermediate reasoning steps before presenting final outcomes, enhancing interpretability and performance~\cite{blackbox, hallucination}. 

In medical applications, 
developing artificial intelligence~(AI) models with reasoning abilities is particularly well-suited, 
as the complexity of medical data and the analytical rigor for accurate diagnosis demand reasoning.
In particular, radiologists often follow a systematic, step-by-step approach to image analysis, adhering to clinical guidelines~\cite{guideline1,guideline2,guideline3}. 
As shown in Fig.~\ref{fig:intro}(a), this involves identifying low-level anomalies, 
integrating patient-specific information ({\em e.g.}, chief complaints), and inferring potential diagnosis. 
These structured reasoning workflows are meticulously documented in clinical reports, 
progressing logically from findings to impressions as dictated by clinical standards. 
Despite this, existing AI research in medical reasoning studies~\cite{mvlmr1, mvlmr2} fails to leverage this structured prior knowledge, and often rely on outcome-based RL methods, neglecting the rich reasoning supervision embedded in clinical reports.

In this paper, we aim to develop a radiology diagnosis model with reasoning capabilities, 
by mining and utilizing process supervision from clinical reports.

For \textbf{dataset construction}, we highlight that daily radiology reports are an invaluable resource for constructing high-quality reasoning chains, as they typically include doctors’ findings and impressions. 
To this end, as illustrated in Fig.~\ref{fig:intro}(b) and Fig.~\ref{fig:reportrminer}(a), 
we propose to construct diverse question-answer pairs based on impressions, and extracts clinically relevant observations in findings as the reasoning processes. We leverage tools like GPT-4o~\cite{gpt4o} to automate the labor-intensive tasks, for example, generating question-answer pairs, extracting clinical observations, and refining them into complete, clinically coherent reasoning chains. Compared to collecting reasoning processes from large visual-language models or relying on manual annotation, extracting reasoning chains directly from textual reports is both scalable and highly reliable.

For \textbf{model development}, we present a generative radiology model with reasoning abilities, 
termed as \textbf{ChestX-Reasoner}. As illustrated in Fig.~\ref{fig:intro}(c), 
we adopt a two-stage training framework that is designed to harness the medical reasoning process supervision~\cite{verystep}. It starts with supervised fine-tuning~(SFT) as a cold start, leveraging both answer-only and reasoning-augmented data for auto-regressive training. 
It then employs reinforcement learning~(RL) with additionally designed process supervision via process reward, that evaluates the correctness of the generated clinical observations compared to the ground-truth report sentences, ensuring a more transparent and reliable training process.

For \textbf{model evaluation}, we curate a reliable reasoning benchmark on Chest X-rays, termed as \textbf{RadRBench-CXR}, that features detailed diagnostic reasoning processes. 
It includes 59K visual question answering~(VQA) samples from three public datasets, namely, MIMIC-CXR~\cite{mimic-cxr}, CheXpert~\cite{chexpert}, and MS-CXR-T~\cite{temporal}. 
The questions span diverse task types that require complex reasoning chains.
Each sample is augmented with reasoning processes mined from its corresponding clinical report, comprising 301k clinically validated reasoning steps. 
As shown in Fig.~\ref{fig:reportrminer}(c), the mined reasoning processes demonstrate high quality upon review by radiologists, and we filtered out unsatisfactory samples to ensure a reliable evaluation. 
Additionally, we present an evaluation metric, \textbf{RadRScore}, that assesses a model's reasoning capabilities across three key dimensions: 
{\em factuality}~(the correctness of generated reasoning), 
{\em completeness}~(the thoroughness in covering clinical findings), 
and {\em effectiveness}~(the necessity and relevance of diagnostic processes).

As a result, our proposed ChestX-Reasoner is a multimodal large language model with medical reasoning for Chest X-ray diagnosis. We compare it against three medical foundation models and five general-domain foundation models in terms of outcome accuracy and reasoning ability, using RadRScore on both RadRBench-CXR and the public CheXbench~\cite{chexagent}. As demonstrated in Fig.~\ref{fig:intro}(d), ChestX-Reasoner achieves significant improvements compared to all baselines. Specifically, it achieves 16\%, 8.5\%, and 18\% improvements in reasoning ability compared to the best medical baseline, the best general baseline, and its base model (Qwen2VL-7B~\cite{qwen2vl}), respectively, as well as 3.3\%, 24\%, and 27\% improvements in outcome accuracy. All code, datasets, and models are open-sourced at \href{https://github.com/MAGIC-AI4Med/ChestX-Reasoner}{ChestX-Reasoner}, and we hope this work inspires further research into developing medical reasoning MLLMs.

\section{Benchmarks}
\label{section:benchmark}
We first present the details for constructing the benchmark, namely, \textbf{RadRBench-CXR}, 
with ground-truth reasoning steps mined from clinical reports. In addition to the final outcome accuracy that measures the correctness of generated answer compared to the ground truth answer, we also provide an evaluation metric to measure the reasoning ability, termed as \textbf{RadRScore}.

\begin{figure}[t!]
    \centering
    \includegraphics[width=1\linewidth]{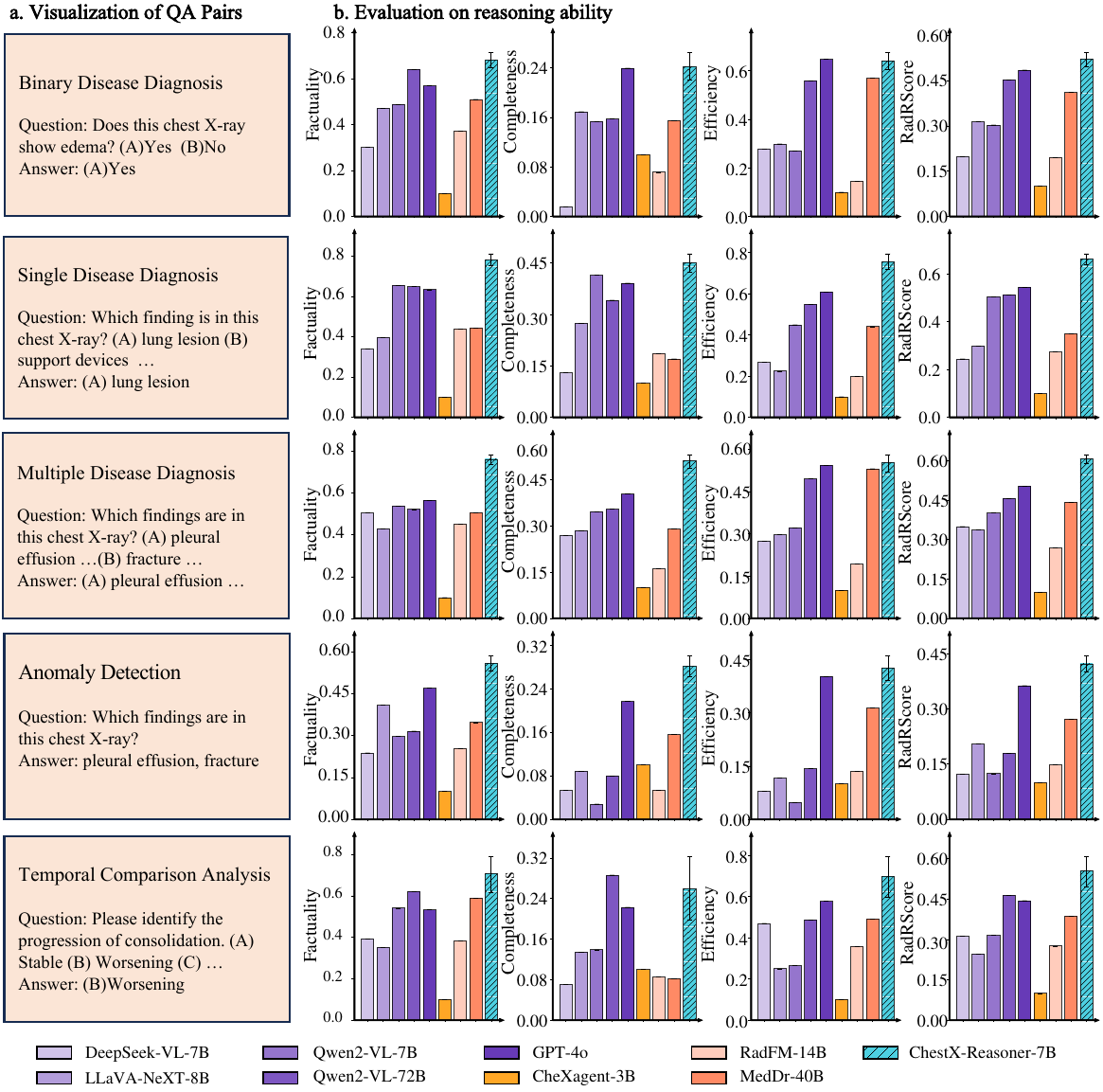}
    \vspace{2pt}
    \caption{\textbf{Evaluation Results of Reasoning Ability.} a. Question-answering pairs visualization of all task types. b. Reasoning abilities on dimensions of factuality, completeness, and effectiveness, as well as the averaged value, RadRScore.}
    \label{fig:reasoning}
\end{figure}

\noindent \textbf{Collecting visual question answering~(VQA) samples.} 
As illustrated in Fig.~\ref{fig:reasoning}(a), 
we collect VQA samples based on the question design framework of CheXbench~\cite{chexagent}, 
which defines comprehensive clinical tasks and generation methods. 
Specifically, we focus on five diagnostic task types:

\begin{itemize}[itemsep=0.3em]
\vspace{-8pt}
    \item \textbf{Binary disease diagnosis:}~yes/no questions about disease presence.
    \item \textbf{Single disease diagnosis:}~multi-choice questions with four options, each representing a disease.
    \item \textbf{Multiple disease diagnosis:}~multi-choice questions with four options, each listing disease combinations, requiring the most clinically relevant answer.
    \item \textbf{Anomaly detection:}~open-ended tasks where models must identify all abnormal findings without predefined options.
    \item \textbf{Temporal comparison analysis:}~disease progression questions using current and prior images, requiring models to determine if a symptom is stable, worsening, or improving.
\vspace{-8pt}
\end{itemize}
The samples are drawn from three datasets—MIMIC-CXR~\cite{mimic-cxr}, CheXpert~\cite{chexpert}, and MS-CXR-T~\cite{temporal}—all paired with clinical reports. To evaluate the performance of our reasoning-enhanced model across multiple clinical centers, we additionally utilize two datasets, RSNA~\cite{rsna} and SIIM~\cite{siim}, that {\em do not} provide associated clinical reports.

\begin{figure*}[ht!]
    \centering    \includegraphics[width=1\linewidth]{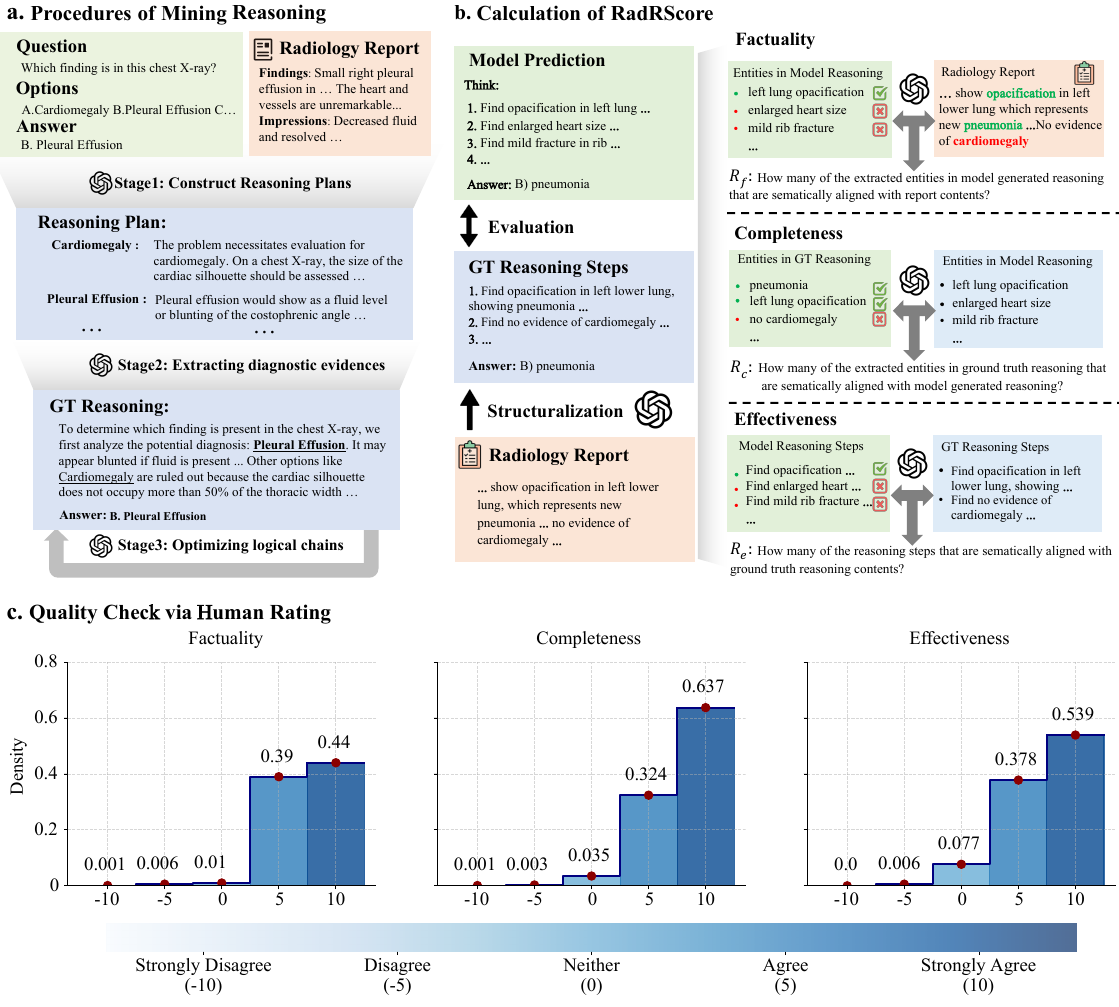}
    \vspace{2pt}
    \caption{\textbf{Reasoning Mining Procedures and RadRScore Calculation} a. Procedures of mining reasoning. We first construct reasoning plans by prompting GPT-4o based on given QA pairs and clinical reports, then extract relative clinical observations from the report as answers for each plan to derive the reasoning steps, and finally, refine the logic to form coherent and complete reasoning chains.
    b. RadRScore calculation targeting on factuality (the correctness of generated reasoning), completeness (the thoroughness in covering clinical findings), and effectiveness (the necessity and relevance of diagnostic processes). 
    c. Human rating on the factuality, completeness, and effectiveness of our mined reasoning processes in the test set, where we use a five-point Likert scale ranging from -10 to 10, based on the provided question, ground truth answer, 
    X-ray images, and clinical report.}
    \label{fig:reportrminer}
\end{figure*}

\paragraph{Mining reasoning chains from reports.} 
As illustrated in Fig.~\ref{fig:intro}(b), and Fig.~\ref{fig:reportrminer}(a), for a subset of VQA samples, we propose to mine the reasoning chains for diagnosis from clinical reports with three key steps:

\begin{itemize}[itemsep=0.3em]
\vspace{-8pt}
    \item \textbf{Constructing reasoning plans:} Given a question-answer pair and the corresponding clinical report, we prompt GPT-4o to generate structured reasoning plans that systematically outline the diagnostic steps leading to the conclusion, for example, the specific symptom or clinical observation. 
    \item \textbf{Extracting diagnostic evidences:} For each reasoning plan, we extract diagnostic evidence from the medical report to construct a single diagnostic reasoning chain. If the report lacks explicit mention of a finding, we infer the response as ‘normal’ or ‘no disease’, given the assumption that physicians do not miss abnormal findings in their reports.
    \item \textbf{Optimizing logical chains:}
    Finally, we integrate all reasoning plans and their extracted answers, refining the logical flow to form a coherent diagnostic reasoning chain.
\vspace{-8pt}
\end{itemize}
To streamline the process, we utilize GPT-4o~\cite{gpt4o} to automate the labor-intensive tasks, 
including constructing reasoning plans, extracting answers from medical reports, 
and refining diagnostic narratives.

\paragraph{Evaluating reasoning with RadRScore.} 
To quantitatively measure the quality of model's intermediate rationales, we deign an automatic evaluation pipeline and a metric, termed as \textbf{RadRScore}. 
As illustrated in Fig.~\ref{fig:reportrminer}(b), 
we first employ large language models~(GPT-4o) to extract entities,
that belong to clinical observations or findings.
For example, from reasoning sentences: 
`Based on the given images, there are no visible signs of lung tissue collapse or an enlarged heart. Besides, there are no signs of pneumothorax...', we can extract entity set $\{$no lung tissue collapse, no enlarged heart, no pneumothorax$\}$.

We use $\text{obs}_{\text{model}}$, $\text{obs}_{\text{gt}}$, and $\text{obs}_{\text{report}}$ to respectively denote the extracted entity set from the model's output, ground-truth reasoning, and corresponding report.
Then, we calculate the values from three dimensions, namely factuality, completeness, and effectiveness based on $\text{obs}_{\text{model}}$, $\text{obs}_{\text{gt}}$, and $\text{obs}_{\text{report}}$: 
\vspace{-3pt}
\begin{itemize}[itemsep = 0.3em]
    \item \textbf{Factuality~($R_f$):} 
    It is calculated as the proportion of elements in $\text{obs}_{\text{model}}$ that are semantically matching with report via GPT-4o. 
    This dimension assesses the factual accuracy of the model's reasoning, ensuring no hallucinated or medically invalid content is included. It can be formulated as:
    \begin{equation}
        R_f=\frac{\text{obs}_{\text{model}} \cap \text{obs}_{\text{report}}}{|\text{obs}_{\text{model}}|}.
    \end{equation}
    \item \textbf{Completeness~($R_c$):} It is calculated as the proportion of elements in $\text{obs}_{\text{gt}}$ that are semantically matching with $\text{obs}_{\text{model}}$, evaluating how comprehensive the model's reasoning is. It can be formulated as:
    \begin{equation}
        R_c=\frac{\text{obs}_{\text{gt}} \cap \text{obs}_{\text{model}}}{|\text{obs}_{\text{gt}}|}.
    \end{equation}
    \item \textbf{Effectiveness~($R_e$):} It is calculated as the proportion of reasoning steps that are semantically included in the ground truth reasoning. Since ground truth reasoning is concise and comprehensive for reaching the final answer, a predicted step is considered effective if it matches one of these steps. It evaluates whether each reasoning step advances the solution in a meaningful way. It can be formulated as:
    \begin{equation}
        R_e=\frac{\text{obs}_{\text{model}} \cap \text{obs}_{\text{gt}}}{|\text{obs}_{\text{model}}|}.
    \end{equation}
\end{itemize}
Notably, when calculating $R_f$, if an element in $\text{obs}_{\text{model}}$ is not found in $\text{obs}_{\text{report}}$, and its meaning contains either `normal' or `no disease', we consider it to be correct, since we assume the physicians may overlook normal observations but typically do not miss abnormal findings in their reports.
Finally, RadRScore is then calculated as the mean value of the three dimensions: 
\begin{equation}
    \text{RadRScore}=(R_f+R_c+R_e)/3
\end{equation}

\paragraph{RadRBench-CXR compilation.}
Leveraging the generated VQA samples, mined reasoning chains, and evaluation metrics, we construct the first visual-language reasoning benchmark, \textbf{RadRBench-CXR}.

To create a balanced and unbiased dataset, we analyze the disease distribution in the original data and adjust the quantities, such that the most frequent disease does not exceed twice the least frequent ones, promoting fairness in the dataset.

Specifically, the benchmark comprises 1.2M answer-only VQA samples for training and 7.3K samples for testing. For a subset of these, we generate 59K reasoning-augmented VQA samples for training and 1K reasoning-augmented samples for testing. We apply strict quality control to the VQA samples with reasoning chains. Samples with a factuality score $R_f$ lower than 1 are filtered out, ensuring that only high-quality, factual reasoning chains are retained in our benchmark. 
Notably, prior to factuality filtering, the generated reasoning processes achieve a significantly higher factuality score ($R_f$) of 0.82, compared to GPT-4o (0.60) and Qwen2VL-72B (0.59), demonstrating the effectiveness of our method to produce high quality reasoning samples.
Additionally, to ensure the reliability of the reasoning evaluation, we conducted human verification, where the radiologists were provided with Chest X-ray images, corresponding reports, and the reasoning outputs generated by our method on the test samples. 
As shown in Fig.~\ref{fig:reportrminer}(c), they achieve average scores of factuality: 7.83, validity: 7.25, and completeness: 7.97 (on a scale from -10 to 10), showing sufficiently high quality. For the released test samples, any reasoning process with a score below 5 in any of the three dimensions was filtered out.
The detailed cases of instruction design, and mined reasoning processes are shown in Supplementary~\ref{app:mined}.

\section{Results: Reasoning Enhanced Medical VLM (ChestX-Reasoner)}
In this section, we compare to various models on the proposed RadRBench-CXR benchmark, emphasizing the superiority of our proposed model over existing multimodal large language models~(MLLMs) in terms of enhanced reasoning ability and outcome accuracy. 

The comparison includes three medical MLLMs—RadFM-14B~\cite{radfm}, CheXagent-3B~\cite{chexagent}, and MedDr-40B~\cite{meddr}, 
and five general MLLMs: DeepSeek-VL-7B~\cite{deepseekvl}, LLaVA-NeXT-8B~\cite{llavanext}, Qwen2VL-7B, Qwen2VL-72B~\cite{qwen2vl}, 
and GPT-4o~\cite{gpt4o}.
The evaluation encompasses the five tasks explained in Sec.~\ref{section:benchmark}, namely, binary disease diagnosis, 
single disease diagnosis, multiple disease diagnosis, anomaly detection,
and temporal comparison analysis.
Due to the lack of clinical reports in SIIM \cite{siim} and RSNA \cite{rsna}, no mined reasoning processes or process supervision were applied to samples from these datasets during our training pipeline.
We use these two datasets for cross-center validation.
Additionally, we provide an ablation study on the impact of different training strategies, underscoring the critical role of process supervision in achieving superior outcomes.
We also provide visualizations of four cases in Supplementary Figs.~\ref{appfig:case1}, \ref{appfig:case2}, \ref{appfig:case3}, and \ref{appfig:case4}, including the VQA questions, extracted reasoning processes, 
and model outputs from our ChestX-Reasoner, GPT-4o, and Qwen2VL-72B. 
Please refer to these visualizations to better understand the effectiveness of our model.

\subsection{Comparison on Reasoning Ability}

While evaluating the reasoning ability of ChestX-Reasoner on the five tasks~(Fig.~\ref{fig:reasoning}(a)) sourced from MIMIC-CXR~\cite{mimic-cxr}, CheXpert~\cite{chexpert}, and MS-CXR-T~\cite{temporal}, we used chain-of-thoughts (CoT) prompt to elicit reasoning from MLLMs and evaluated each model with RadRScore, measuring factuality, completeness, and effectiveness of the reasoning.

ChestX-Reasoner~(7B) consistently outperformed all other MLLMs across tasks, showing an average RadRScore improvement of 18\% over its base model, Qwen2VL-7B (from 0.349 to 0.531; see last column of Fig.~\ref{fig:reasoning}(b)). 
Among general domain models, GPT-4o and Qwen2VL-72B achieved RadRScores of 0.472 and 0.407, respectively. The best medical domain baseline, MedDr-40B, reached 0.367. Although CheXagent-3B produced competing outcome accuracy compared to ours, its reasoning scores were near zero across all tasks.

\textbf{Factuality.} 
As shown in the first column of Fig.~\ref{fig:reasoning}(b), 
ChestX-Reasoner achieved the following scores:
0.682 \;(95\%CI\;=\;0.646–0.710) on binary disease diagnosis,
0.751 \;(95\%CI\;=\;0.722–0.782) on single disease diagnosis,
0.778 \;(95\%CI\;=\;0.753–0.800) on multiple disease diagnosis,
0.588 \;(95\%CI\;=\;0.560–0.615) on anomaly detection, and
0.706 \;(95\%CI\;=\;0.617–0.790) on temporal comparison analysis. 
These results outperform the best general MLLM, GPT-4o with average scores of 0.534, 0.617, 0.554, 0.460, and 0.532 on the five tasks, respectively.
Additionally, ChestX-Reasoner also outperforms the best medical MLLM, MedDr-40B, which achieved average scores of 0.570, 0.633, 0.567, 0.471, and 0.532 on these tasks. These findings demonstrate that ChestX-Reasoner exhibits superior clinical factual correctness compared to other baselines.

\textbf{Completeness.} 
As shown in the second column of Fig.~\ref{fig:reasoning}(b), 
ChestX-Reasoner achieved the following scores:
0.240 \;(95\%CI\;=\;0.218–0.262) on binary disease diagnosis,
0.430 \;(95\%CI\;=\;0.403–0.458) on single disease diagnosis,
0.520 \;(95\%CI\;=\;0.497–0.541) on multiple disease diagnosis,
0.282 \;(95\%CI\;=\;0.262–0.301) on anomaly detection, and
0.259 \;(95\%CI\;=\;0.199–0.323) on temporal comparison analysis.
Among the baselines, general-domain MLLMs like GPT-4o and Qwen2VL-72B achieved comparable performance, except for the anomaly detection task.
Qwen2VL-72B failed in anomaly detection due to the lack of given options.
In contrast, medical-domain MLLMs performed significantly worse in this dimension. The best-performing medical MLLM, MedDr-40B, achieved completeness scores of only 0.155 on binary disease diagnosis, 
0.171 on single disease diagnosis, 0.291 on multiple disease diagnosis, 0.156 on anomaly detection, and 0.082 on temporal comparison analysis.
These results underscore the superior thoroughness of ChestX-Reasoner in covering clinical findings while highlighting the limitations of current medical-domain MLLMs.

\textbf{Effectiveness.} 
As shown in the third column of Fig.~\ref{fig:reasoning}(b), 
ChestX-Reasoner achieved the following scores:
0.661 \;(95\%CI\;=\;0.627–0.695) on binary disease diagnosis,
0.688 \;(95\%CI\;=\;0.649–0.727) on single disease diagnosis,
0.597 \;(95\%CI\;=\;0.565–0.627) on multiple disease diagnosis,
0.415 \;(95\%CI\;=\;0.380-0.450) on anomaly detection, and
0.698 \;(95\%CI\;=\;0.600–0.796) on temporal comparison analysis.
In this dimension, the medical-domain MLLM, MedDr-40B, achieves performance comparable to the best general-domain MLLM, GPT-4o, but still falls significantly short of ChestX-Reasoner.
Our ChestX-Reasoner outperforms both GPT-4o and MedDr-40B, with improvements of 5.8\% and 34.0\%, respectively. These results underscore the superior necessity and relevance of the diagnostic processes generated by our model.

\begin{figure}[t!]
    \centering
    \includegraphics[width=0.98\linewidth]{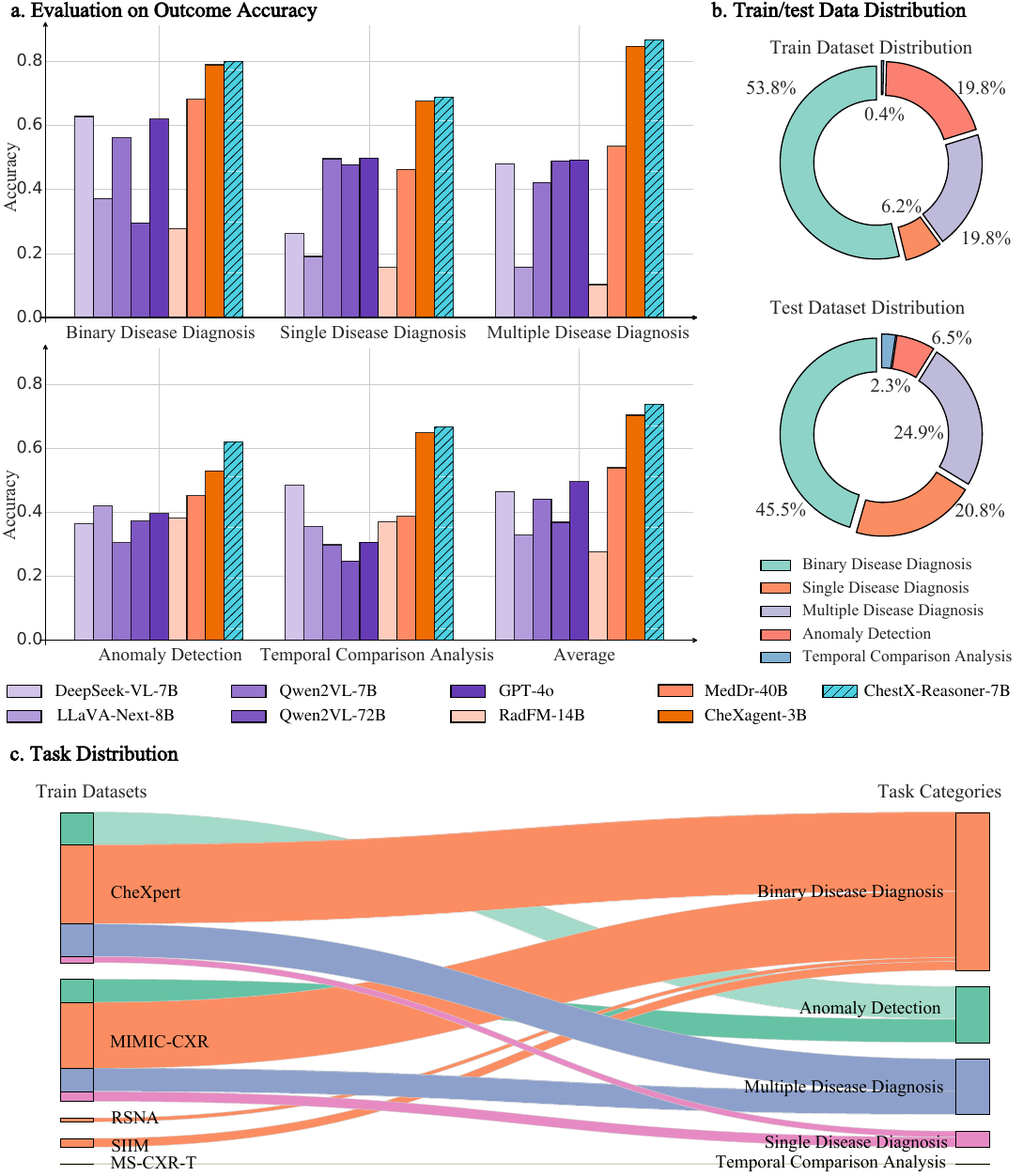}
    \caption{\textbf{Evaluation Results of Outcome Accuracy and Data Statistics.} a. Outcome accuracy evaluated on five tasks. b. Train and test data distribution from different data sources. c. Task distribution from different data sources.}
    \label{fig:correctness}
\end{figure}

\subsection{Comparison on Outcome Accuracy}
Next, we evaluated the ability of ChestX-Reasoner to output correct outcome after generating reasoning processes on the same five tasks~(Fig. \ref{fig:correctness}(b)) sourced from MIMIC-CXR~\cite{mimic-cxr}, CheXpert~\cite{chexpert}, MS-CXR-T~\cite{temporal}, SIIM~\cite{siim}, and RSNA~\cite{rsna} as shown in Fig. \ref{fig:correctness}(c).
%
Since anomaly detection is an open-ended generation task, 
we adopt RaTEscore~\cite{ratescore} for evaluation, 
while for other close-ended tasks with provided options, we directly verify whether the generated answer is the same with the ground truth label.
Results demonstrated that ChestX-Reasoner achieves superior performance compared to all baselines with 3\% overall improvements compared to the existing state-of-the-art MLLM~(CheXagent-3B), which was trained on significantly more data (8.5M training samples compared to 1.2M samples used in ChestX-Reasoner).
Additionally, ChestX-Reasoner outperforms its base model~(Qwen2VL-7B) with 27\% overall improvements, and achieves 24\% averaged improvements comapred to best general domain MLLM GPT-4o.

\textbf{Disease diagnosis.} 
For all the disease diagnosis tasks, our model achieves comparable performance to existing state-of-the-art model~(CheXagent-3B), with accuracy of 0.800 \;(95\%CI\;=\;0.776–0.821) on binary disease diagnosis,
0.688 \;(95\%CI\;=\;0.641–0.732) on single disease diagnosis,
and 0.785 \;(95\%CI\;=\;0.735–0.831) on multiple disease diagnosis.
Among general domain MLLM, GPT-4o achieves the best performance with accuracy of 0.620 on binary disease diagnosis, 0.498 on single disease diagnosis, and 0.492 on multiple disease diagnosis, comparable to MedDr-40B, but much worse than our ChestX-Reasoner. This demonstrate the superior performance of our model on solving the disease diagnosis tasks.
 
\textbf{Anomaly detection.} 
Here, our model greatly improves CheXagent-3B, 
with RaTEscore of 0.621 (95\% CI = 0.608 – 0.633) compared to 0.529, achieving 17\% improvements.
General domain MLLMs like GPT-4o and Qwen2VL-72B achieves RaTEscore of 0.396 and 0.372, which is even worse than the LLaVA-NeXT-8B of 0.419.
This comparison shows that ChestX-Reasoner also performs superior on open-ended generation task compared to all other baselines.

\textbf{Temporal comparison analysis.} 
Our model achieves comparable performance compared to CheXagent-3B, with an accuracy of 0.661 \;(95\%CI\;=\;0.537–0.778).
Among general MLLMs and other medical MLLMs, DeepSeek-VL-7B and MedDr-40B perform best, with an accuracy of 0.484 and 0.387, which remains inferior to our model.

\begin{figure}[t!]
    \centering
    \includegraphics[width=0.95\linewidth]{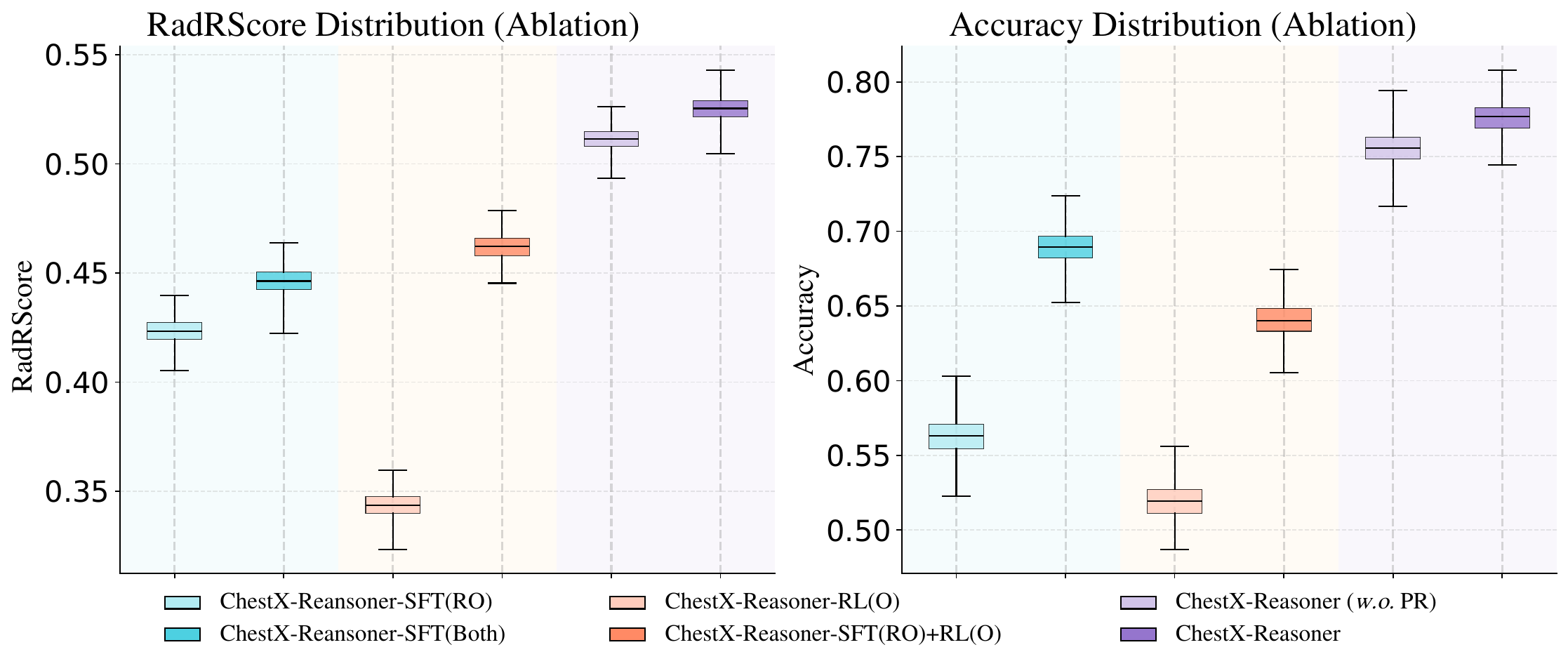}
    \vspace{3pt}
    \caption{Ablation studies on both reasoning ability and outcome accuracy of our trained models including ChestX-Reasoner-SFT(RO), ChestX-Reasoner-SFT(Both), ChestX-Reasoner-RL(O), ChestX-Reasoner-SFT(RO)+RL(O), ChestX-Reasoner ({\em w.o.} Process Reward), and the final ChestX-Reasoner.}
    \label{fig:ablation}
\end{figure}

\subsection{Ablation on Training Strategy}

Recent reasoning-enhanced training strategies, prevalent in mathematics and programming, include supervised fine-tuning (SFT) on reasoning samples, outcome-based reinforcement learning (RL) on answer-only samples with chain-of-thoughts (CoT) prompt, and DeepSeek-R1’s two-stage approach (SFT followed by RL). However, these methods are largely unexplored for medical especially radiology MLLMs. 

To bridge this gap, we conducted an ablation study~(Fig.~\ref{fig:ablation}) comparing various strategies, resulting in models: 
\begin{itemize}[itemsep=0.3em]
\vspace{-8pt}

    \item \textbf{ChestX-Reasoner-SFT(RO)}: Model trained on only reasoning-augmented samples with SFT.
    \item \textbf{ChestX-Reasoner-SFT(Both)}: Model trained on both reasoning-augmented samples and answer-only samples with SFT.
    \item \textbf{ChestX-Reasoner-RL(O)}: Model trained on answer-only samples with outcome-based RL.
    \item \textbf{ChestX-Reasoner-SFT(RO)+RL(O)}: Model trained on reasoning-augmented samples with SFT, followed by outcome-based RL on answer-only samples.
    \item \textbf{ChestX-Reasoner ({\em w.o.}~PR)}: Model trained on both reasoning-augmented and answer-only samples with SFT, followed by outcome-based RL on answer-only samples.
    \item \textbf{ChestX-Reasoner}: Our final model, trained on both reasoning-augmented and answer-only samples with SFT followed by RL on both types of samples with additional process reward. See details in Sec.~\ref{sec:method}.
\end{itemize}
Notably, reasoning-augmented samples are data with clinical report and mined reasoning sentences, while answer-only samples are data without mined reasoning and clinical reports.

\textbf{SFT and RL are both necessary.} 
As shown in Fig.~\ref{fig:ablation}, models trained solely with outcome-based RL~(ChestX-Reasoner-RL(O), averaged RadRScore: 0.343, 95\% CI: 0.347–0.362; 
outcome accuracy: 0.471, 95\% CI: 0.452–0.491) or with SFT on reasoning-augmented samples alone~(ChestX-Reasoner-SFT(RO), averaged RadRScore: 0.423, 95\% CI: 0.412–0.434; outcome accuracy: 0.566, 95\% CI: 0.545–0.586) both perform poorly. 
SFT alone is limited by its reliance on exact ground-truth replication, requiring extensive reasoning processes, while RL alone lacks necessary reasoning knowledge in pre-trained base MLLMs. These findings indicate that combining SFT and RL is crucial for developing effective medical reasoning MLLMs.

\textbf{Answer supervision is helpful for domain alignment before enhancing reasoning.} 
As shown in Fig.~\ref{fig:ablation}, ChestX-Reasoner-SFT(Both) achieves 5\% improvements in reasoning and 19.9\% improvements in outcome accuracy compared to ChestX-Reasoner-SFT(RO).
Additionally, starting from ChestX-Reasoner-SFT(Both), ChestX-Reasoner~({\em w.o.}~process reward) achieves 14.5\% improvements in reasoning and 9.5\% improvements in outcome accuracy over ChestX-Reasoner-SFT(RO)+RL(O), which begins from ChestX-Reasoner-SFT(RO).
These results demonstrate that leveraging extensive answer supervision from answer-only samples during SFT is beneficial for medical domain alignment, thereby producing better reasoning MLLMs in subsequent steps.

\textbf{Process supervision is essential for reasoning ability.} 
Adding process supervision further enhances reasoning ability. 
ChestX-Reasoner achieves an averaged RadRScore of 0.525 (95\% CI: 0.515–0.536), a 2.5\% improvement over ChestX-Reasoner~({\em w.o.}~process reward; 0.512, 95\% CI: 0.501–0.513). The enhanced reasoning ability also translates into improved outcome accuracy. These results underscore the importance of process supervision in boosting the reasoning performance of medical MLLMs, ultimately leading to better diagnostic accuracy.

\section{Discussion}
This study systematically investigates the development of medical reasoning-enhanced multimodal foundation models. Unlike general computer vision, where visual reasoning is often implicit~\cite{vlmr1,vlmr2}, medical imaging analysis involves explicit, standardized reasoning workflows. Radiologists routinely document detailed reasoning in clinical reports, logically progressing from findings to diagnostic conclusions—offering valuable process supervision signals. Based on this insight, our main contributions are threefold:

\textbf{A scalable reasoning mining pipeline using clinical reports.} 
Unlike prior approaches that rely solely on chain-of-thoughts distillation~\cite{medxpertqa,huatuoo1,videogpt}, our pipeline extracts reasoning cases directly from manually written clinical reports, ensuring higher factual accuracy and verifiability. While demonstrated here with Chest X-ray reports, this method can be readily extended to other radiology modalities, providing a scalable and cost-effective source of validated reasoning samples for clinical AI development.

\textbf{ChestX-Reasoner, a process-supervised radiology foundation model with reasoning.} 
We introduce ChestX-Reasoner, a multimodal large language model trained with stepwise process supervision via process reward to mirror the structured reasoning of clinical diagnostics. ChestX-Reasoner achieves state-of-the-art performance in both reasoning and outcome accuracy on Chest X-ray diagnostic tasks, outperforming general models (DeepSeek-VL-7B~\cite{deepseekvl}, LLaVA-NeXT-8B~\cite{llavanext}, GPT-4o~\cite{gpt4o}, Qwen2VL-7B~\cite{qwen2vl}, and Qwen2VL-72B~\cite{qwen2vl}) and specialized medical models (RadFM-14B~\cite{radfm}, MedDr-40B~\cite{meddr}, CheXagent-3B~\cite{chexagent}) across all five benchmark tasks.

\textbf{RadRBench-CXR, a reliable reasoning-focused radiology benchmark.} 
For evaluation, we introduce RadRBench-CXR, derived from real clinical reports and designed to assess both outcomes and reasoning quality. Beyond outcome accuracy, our RadRScore comprehensively evaluates generated reasoning in terms of factuality, completeness, and effectiveness, providing a robust metric for benchmarking medical MLLMs.

\vspace{5pt}
Based on our experimental results, we derive the following key findings:

\textbf{LLMs can transform clinical reports into reasoning samples across various radiology tasks.} The development of medical reasoning MLLMs is often hindered by the scarcity and high cost of high-quality, reliable multimodal reasoning datasets.
In this work, we show that existing clinical reports, combined with effective prompt engineering, can enable LLMs to automatically generate diverse, well-structured reasoning-labeled training instances across a wide range of radiology tasks. 
By mining reasoning processes from clinical reports, our generated reasoning achieves a significantly higher factuality score-defined by RadRScore, which measures the correctness of generated reasoning-reaching 0.82, compared to GPT-4o (0.60) and QwenVL-72B (0.59). Additionally, human evaluations conducted by radiologists (Fig.~\ref{fig:reportrminer}(c)) reveal strong satisfaction with the generated reasoning, with average scores of factuality: 7.83, validity: 7.25, and completeness: 7.97 (on a scale from -10 to 10). These results highlight the potential of leveraging LLMs to convert clinical reports into high-quality reasoning samples, offering promising implications for broader applications in medical AI and beyond.

\textbf{Enhancing the initial medical capabilities of base MLLMs via SFT as a cold-start is necessary.}
General-domain base MLLMs often lack the specialized medical knowledge required for clinical reasoning tasks. 
Our experiments reveal that directly applying reinforcement learning (RL) to such models, {\em e.g.}, {ChestX-Reasoner-RL(O)}, results in poor performance and unstable optimization. This underscores the necessity of initializing general-domain MLLMs with medical knowledge through domain-specific pretraining or task-adaptive supervised fine-tuning before applying RL. Without this initialization, models are prone to convergence failures and cannot reliably develop clinical reasoning skills.

\textbf{Process rewards are critical for RL in developing medical reasoning MLLMs.}
While recent advances in general domain reasoning MLLMs have inspired medical MLLMs~\cite{mvlmr1, mvlmr2} to adopt outcome-based RL strategies, these rely on \textit{self-exploration} without explicit reasoning supervision. However, due to the complexity and domain-specificity of medical reasoning, this approach proves unreliable. 
In this paper, we demonstrate that process rewards are critical in the medical imaging domain, significantly enhancing the quality of the reasoning process. Our model, ChestX-Reasoner, achieves notable improvements in both reasoning ability and outcome accuracy compared to its counterpart without process supervision, as shown in Fig.~\ref{fig:ablation}.

These findings can inspire future research in key areas, including the development of robust and generalizable medical reasoning foundation models, scalable generation of clinically validated reasoning data, and the adaptation of process supervision techniques to enhance medical reasoning generation.

From a clinical perspective, our process-supervised medical reasoning model, ChestX-Reasoner, offers significant advantages over prior outcome-only models. By explicitly demonstrating step-by-step imaging interpretation, it produces outputs that are more interpretable to clinicians and better aligned with clinical workflows. This enables transparent decision-support evidence in radiology tasks such as reporting, differential diagnosis, and multidisciplinary case discussions. Additionally, the improved factual consistency and reasoning traceability enhance auditability, reducing challenges in human-AI interaction within clinical practice. As a proof of concept, ChestX-Reasoner highlights a more effective approach to developing reasoning-enhanced medical foundation models, paving the way for integration into real-world clinical systems where explainability, reliability, and safety are essential.

Lastly, we have to emphasize that our study presents several \textbf{limitations}, which highlight the opportunities for future work.
First, we focus on Chest X-rays to demonstrate our approach due to the availability of public datasets and computational constraints. 
However, our methods are generalizable to other imaging modalities with detailed reports ({\em e.g.}, CT, MRI, ultrasound), making their extension a key direction for future work. 
Additionally, our approach builds on the general-domain vision-language model Qwen2VL-7B~\cite{qwen2vl}, which serves as a strong baseline. Future research could explore larger models or those with medical domain-specific pretraining. Notably, our framework is model-agnostic and can be easily adapted to other base models.
Second, {ChestX-Reasoner} currently relies on reasoning-augmented samples and rule-based factuality scores for process supervision. As our primary goal is to emphasize the value of process-level rewards, refining the supervision mechanism is left for future work. We aim to explore more robust strategies, such as developing medical-specific agentic reward models~\cite{agentic1}, to better manage samples without annotated reasoning references.

\begin{figure*}[t!]
    \centering
    \includegraphics[width=1\linewidth]{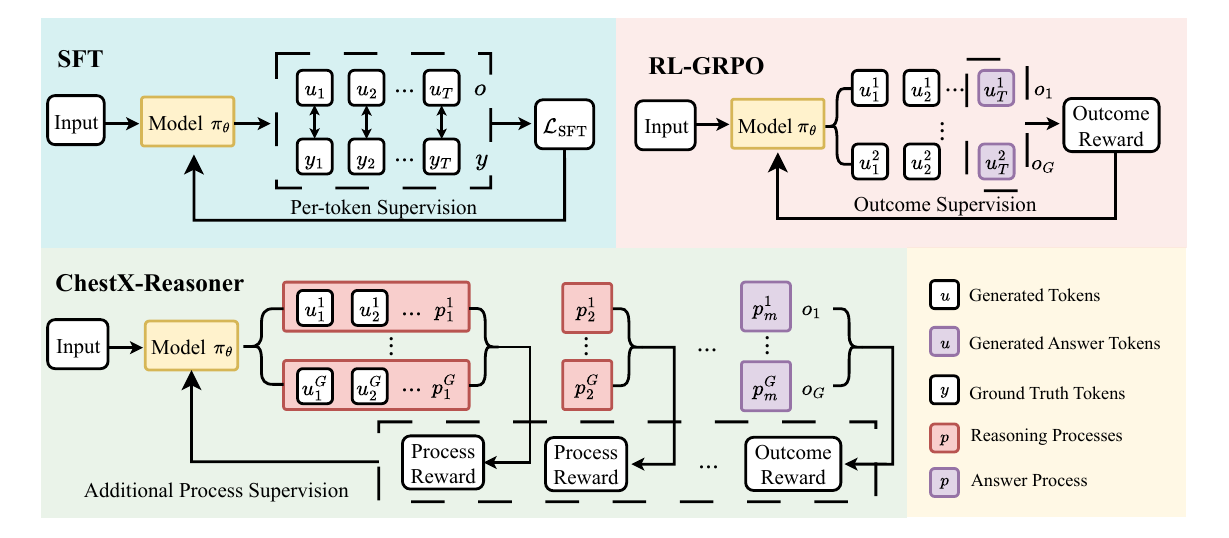}
    \vspace{2pt}
    \caption{\textbf{Illustration of different supervision methods.} The top left illustrates the per-token supervision of supervised fine-tuning. The top right shows the outcome supervision on only several tokens related to answer or format. The bottom indicates our process supervision on every several tokens that contributes to a process.}
    \label{fig:method}                                      
\end{figure*}
\section{Methods}\label{sec:method}

In this section, we first introduce the necessary preliminaries and then outline the training procedure for ChestX-Reasoner, which starts with supervised fine-tuning on both answer-only and reasoning-augmented data, followed by reinforcement learning with additional process rewards.
We adopt Qwen2VL-7B~\cite{qwen2vl} as our base model for its strong reasoning performance in the general vision-language domains~\cite{qwenbase1,qwenbase2,qwenbase3}. Details of the training hyper-parameters are provided in the Supplementary~\ref{app:training}.

\subsection{Preliminaries} 
\paragraph{Problem statement.} Assuming dataset $\mathcal{D}=\{(\mathcal{X}_{i},\mathcal{Q}_{i},\mathcal{P}_i,\mathcal{C}_i,\mathcal{A}_{i})\}_{i=1}^{N}$, where $N$ denotes the number of samples, $\mathcal{X}_i$ represents the input images, $\mathcal{Q}_i$ denotes the associated text instruction, $\mathcal{P}_i$ refers to the mined reasoning processes, $\mathcal{C}_i$ is the corresponding clinical report, and $\mathcal{A}_i$ is the ground-truth answer.
Our objective is to develop a multimodal large language model~(MLLM) that enhances Chest X-ray interpretation by improving both reasoning capability and outcome accuracy on $\mathcal{D}$. Formally, the objective is to generate the desired output sentences given the input images $\mathcal{X}$ and the text instruction $\mathcal{Q}$:
\begin{equation}
    \mathcal{Y}=\pi_{\theta}(\cdot ;\mathcal{X},\mathcal{Q}),
\end{equation}
where $\pi_{\theta}$ is the policy model parameterized by $\theta$, and $\mathcal{Y}$ denotes the output sentences. Notably, there are samples without clinical report and mined reasoning processes in $\mathcal{D}$, which means 
their $\mathcal{P}=\emptyset$, and $\mathcal{C}=\emptyset$. We distinguish them by dividing $\mathcal{D}=\{\mathcal{D}_R, \mathcal{D}_A\}$ with reasoning-augmented data~($\mathcal{D}_R$), 
and answer-only data~($\mathcal{D}_A$):
\begin{itemize}[itemsep=0.3em]
\vspace{-8pt}
    \item \textbf{Reasoning-augmented data}~($\mathcal{D}_R$): 
    set of samples with clinical report and mined reasoning sentences.
    \item \textbf{Answer-only data}~($\mathcal{D}_A$): 
    set of samples without ground-truth reasoning and clinical reports, 
    which means their $\mathcal{P}=\emptyset$, and $\mathcal{C}=\emptyset$.
\end{itemize}

To enable the model to function in both reasoning and direct response modes, we design two instruction formats for each sample: one with a chain-of-thoughts (CoT) prompt that elicits reasoning, and one without, which directly requests the answer. Details of these instruction formats are provided in Supplementary~\ref{app:instruction}. By leveraging both CoT and non-CoT prompts, our approach allows for disabling intermediate reasoning, thereby substantially reducing inference time when reasoning is unnecessary.

\vspace{5pt}\noindent \textbf{DeepSeek-R0 with outcome rewards and GRPO algorithm.} Recent medical studies~\cite{mvlmr1,mvlmr2} on enhancing reasoning abilities have adopted the training procedures of DeepSeek-R0~\cite{deepseekr1}, which aims to guide the model for reasoning with reinforcement learning on answer-only samples~($\mathcal{D}_A$). The optimization algorithm employed is Group Relative Policy Optimization (GRPO)~\cite{grpo} as RL-GRPO illustrated in Fig.~\ref{fig:method}. 
Compared to Policy Gradient Optimization~(PPO)~\cite{ppo} algorithm, GRPO eliminates the need for a separate critic model and estimates baseline based on group scores, thereby significantly lowering computational overhead.
With regard to the reward functions used in the algorithm, they are based on the outcome of model output, namely \texttt{outcome format} reward and \texttt{outcome accuracy} reward.
Outcome format reward enables the output to be more structured, thus facilitating further visualization, and performance evaluation. Outcome accuracy reward is the evaluation metric between output answer and the ground truth answer. They evaluate the overall output while overlooking the specific characteristics of individual processes. For comparison, we trained ChestX-Reasoner-RL(O) following this pipeline.

\vspace{5pt}\noindent \textbf{DeepSeek-R1.}
In the DeepSeek-R1 model~\cite{deepseekr1}, an additional supervised fine-tuning step as SFT shown in Fig~\ref{fig:method} is utilized as a cold-start initialization before reinforcement learning.
This involves auto-regressive training of the base model on samples with reasoning chains ($\mathcal{D}_R$). The objective maximizes the likelihood of generating both reasoning processes and answers given input $\mathcal{X}$, and $\mathcal{Q}$:
\begin{equation}
\mathcal{L}_{\text{SFT}}=-\mathbb{E}_{(\mathcal{X},\mathcal{Q},\mathcal{P},\mathcal{C},\mathcal{A})\sim\mathcal{D}_R}\sum_{t=1}^{T} \log\pi_{\theta}(y_{t}\mid \mathcal{X},\mathcal{Q},y_{<t}),
\label{eq:sft}
\end{equation}
where $y$ is the concatenated sequence tokens of mined reasoning processes $\mathcal{P}$ and ground truth answer $\mathcal{A}$, $t$ is the index of decoded tokens $y$, and $T$ is the total number of supervised tokens. 
For comparison, we trained ChestX-Reasoner-SFT(RO)+RL(O) following this pipeline.

\subsection{ChestX-Reasoner}
After introducing the preliminaries, we now detail the training pipeline for ChestX-Reasoner in the following.

\paragraph{Stage-I: supervised fine-tuning.} 
Different to Deepseek-R1, we adopt auto-regressive training of base model on both $\mathcal{D}_A$, and $\mathcal{D}_R$.
The training maximizes the following objective:
\begin{equation}
\mathcal{L}_{\text{SFT}}=-\mathbb{E}_{(\mathcal{X},\mathcal{Q},\mathcal{P},\mathcal{C},\mathcal{A})\sim\{\mathcal{D}_A,\mathcal{D}_R\}}\sum_{t=1}^{T} \log\pi_{\theta}(y_{t}\mid \mathcal{X},\mathcal{Q},y_{<t}).
\label{eq:sft}
\end{equation}
We find that tuning the model exclusively on either $\mathcal{D}_A$ or $\mathcal{D}_R$ fails to achieve the desired objective. Specifically, solely using the final answer as supervision, without incorporating intermediate reasoning steps, leads to overfitting to the answer patterns and an undesirable loss of reasoning ability. Conversely, tuning the base model on the small-scale reasoning-augmented samples alone fails to leverage the valuable labels available in the full dataset, resulting in suboptimal model performance and the inability to directly output the answer.

\paragraph{Stage-II: reinforcement learning with process reward.} With SFT model as an initialization, we then employ GRPO algorithm to perform reinforcement learning with additional process reward as ChestX-Reasoner shown in Fig.~\ref{fig:method}.
Specifically, for each input $ (\mathcal{X},\mathcal{Q},\mathcal{P},\mathcal{C},\mathcal{A})\in \mathcal{D} $, GRPO samples a group of output sentences $\mathcal{O}\{o_1, o_2, \dots, o_G\}$ from the old model \(\pi_{\text{old}}\), where $G$ is the pre-defined sentence number in a group.
It then optimizes the model \(\pi_\theta\) by maximizing the following objective function:
\begin{equation}
    \small
J_{\text{GRPO}} = \mathbb{E}_{(\mathcal{X},\mathcal{Q},\mathcal{P},\mathcal{C},\mathcal{A}) \in \mathcal{D}, \{o_i\} \sim \pi_{\text{old}}(\mathcal{O}| \mathcal{X},\mathcal{Q})} \left[ \frac{1}{G} \sum_{i=1}^{G} \left( \min \left( \frac{\pi_\theta(o_i | \mathcal{X},\mathcal{Q})}{\pi_{\text{old}}(o_i | \mathcal{X},\mathcal{Q})}, 1 - \epsilon, 1 + \epsilon \right) A_i - \beta \, D_{\text{KL}} \left( \pi_\text{old}\parallel \pi_\theta \right) \right) \right],
\end{equation}
where \(\epsilon\) and \(\beta\) are hyperparameters controlling the clipping range and the strength of the KL divergence penalty, respectively. 
KL penalty constrains the updating calculated as: $D_{\text{KL}} \left( \pi_{_{\text{old}}} \parallel \pi_\theta \right) = \log \left( \frac{\pi_{{\text{old}}}(o_i | \mathcal{X},\mathcal{Q})}{\pi_{{\theta}}(o_i |\mathcal{X}, \mathcal{Q})} \right).$
The advantage \(A_i\) is computed using a group of rewards \(\{r_1, r_2, \dots, r_G\}\) corresponding to the group outputs calculated as
$A_i = \frac{r_i - \text{mean}(\{r_1, r_2, \dots, r_G\})}{\text{std}(\{r_1, r_2, \dots, r_G\})}.$
Samples from $\mathcal{D}_A$ use the outcome format reward and the outcome accuracy reward, whereas for samples from $\mathcal{D}_R$, an additional process factuality reward is included, which are all defined as follows:
\begin{itemize}[itemsep=0.3em]
\vspace{-8pt}
    \item \textbf{Outcome Format Reward} $R_{\text{format}}$: A reward of 1 is assigned if the model adheres to a specific output structure, specifically <think> </think>, <answer> </answer> of samples in $\mathcal{D}_R$ and <answer> </answer> of samples in $\mathcal{D}_A$; otherwise, the reward is 0.
    \item \textbf{Outcome Accuracy Reward} $R_{\text{oa}}$: 
    An evaluation metric is used to measure the correctness of the model's output in comparison to the ground truth answer. Specifically, we extract the content enclosed within <answer> and </answer> tags as the model's predicted answer. For close-ended questions with provided options, the reward is set to 1 if the predicted answer is the same with the ground truth option; otherwise, it is set to 0. For open-ended generation tasks, we employ RaTEScore~\cite{ratescore} on the extracted answer as the reward. 
    \item \textbf{Process Factuality Reward} $R_f$: 
    The reward is calculated as the factuality score defined in RadRScore, which evaluates the proportion of correctness of the generated clinical observations compared to the ground-truth report sentences.
\vspace{-8pt}
\end{itemize}

The calculation of reward can be formulated as:
\begin{equation}
    \label{eq:init}
    r_i=r(o_i;(\mathcal{X},\mathcal{Q},\mathcal{P},\mathcal{C},\mathcal{A}))=\left\{
    \begin{array}{ll}
    R_{\text{format}}(o_i;\text{format})+R_{\text{oa}}(o_i;\mathcal{A}), & (\mathcal{X},\mathcal{Q},\mathcal{P},\mathcal{C},\mathcal{A})\in \mathcal{D}_A, \\
    R_{\text{format}}(o_i;\text{format})+R_{\text{oa}}(o_i;\mathcal{A})+R_f(o_i;\mathcal{C},\mathcal{A}), & (\mathcal{X},\mathcal{Q},\mathcal{P},\mathcal{C},\mathcal{A})\in\mathcal{D}_R, \\
    \end{array} \right. 
\end{equation}
where $\mathcal{C}$ is the clinical report, and $\mathcal{A}$ is ground truth answer.
The distinction among these rewards lies in their supervision signals.
From the equation, we can easily find that, outcome format reward and outcome accuracy reward are calculated based on only several tokens describing format or final answer compared to a specific format and ground truth answer $\mathcal{A}$.
These two rewards are typically used in many prior work~\cite{deepseekr1,vlmr1,vlmr2,mvlmr1,mvlmr2}, as well as in ChestX-Reasoner-RL(O) and ChestX-Reasoner-SFT(RO)+RL(O), lacking supervision of reasoning processes.
To incentivize high-quality reasoning, we preserves the robust reasoning supervision embedded in clinical reports via process reward, ensuring a more transparent and reliable reasoning process.

\section*{Data availability}
All datasets used in this study are publicly accessible.
This study utilized datasets that are all publicly accessible. 
The RadRBench-CXR evaluation benchmark, and the model weights of all stages and variants are released in \href{https://github.com/MAGIC-AI4Med/ChestX-Reasoner}{ChestX-Reasoner-Code}.

\section*{Code availability}
The code used for experiments in this study is released in \href{https://github.com/MAGIC-AI4Med/ChestX-Reasoner}{ChestX-Reasoner-Code}. 
It includes the code to create reasoning processes from reports and the code to train the model ChestX-Reasoner.
We release the model weights in \href{https://huggingface.co/byrLLCC/ChestX-Reasoner}{ChestX-Reasoner-Model} and other variants will be hosted on HuggingFace (https://huggingface.co/) before publication.

\section*{Author Contributions}
Ziqing Fan, Chaoyi Wu, and Weidi Xie designed the training pipeline, and Ziqing Fan carried out the model training.
Cheng Liang, Ziqing Fan, Chaoyi Wu designed the data collection, and Cheng Liang, and Ziqing Fan carried out creating reasoning processes.
Cheng Liang designed and carried out the model evaluation.
All authors contributed to the drafting and revision of the manuscript. 
Weidi Xie, and Yanfeng Wang supervised and guided the research.

\bibliographystyle{unsrt}
\bibliography{main} 
\clearpage
\section{Supplementary}
\captionsetup[figure]{name=Supplementary Figure}
\captionsetup[table]{name=Supplementary Table}
\setcounter{table}{0}
\setcounter{figure}{0}

\begin{table}[ht!]
\centering
  \caption{Training hyper-parameters in training stage 1.}
  \label{apptab:param1}
  \begin{tabular}{ll}
    \hline
    Parameter name & Value \\
    \hline
    Global batchsize & 512 \\
    Learning rate& 2e-6 \\
    Optimizer& AdamW \\
    $\beta_1$ of optimizer& 0.9 \\
    $\beta_2$ of optimizer& 0.95 \\
    Warmup ratio& 0.1 \\
    Learning rate scheduler& Cosine \\
    Gradient clipping threshold& 1.0 \\
    Weight decay& 0.0 \\
    Platform & 8 Tesla A100 GPUs \\
    Training steps & 3500 steps \\
    Training times& about 2 days \\
    DeepSpeed& zero2 \\
    \hline
  \end{tabular}
\end{table}
    
\begin{table}[ht!]
\centering
  \caption{Training hyper-parameters in training stage 2.}
  \label{apptab:param2}
  \begin{tabular}{ll}
    \hline
    Parameter name & Value \\
    \hline
    Global batchsize & 256 \\
    Updating batchsize & 256 \\
    KL loss coefficient & 0.0001 \\
    Gradient clipping threshold& 1.0 \\
    Entropy loss coefficient&0.0001 \\
    Group number~($G$)& 8 \\
    Learning rate& 5e-7 \\
    Optimizer& AdamW \\
    $\beta_1$ of optimizer& 0.9 \\
    $\beta_2$ of optimizer& 0.95 \\
    Platform & 8 Tesla A100 GPUs \\
    Training steps& 400 steps \\
    Training times& about 3.5 days \\
    Distributed platform&  FSDP without offloading\\
    Reward function & outcome format and outcome accuracy, and process factuality rewards \\ 
        \hline
  \end{tabular}
\end{table}

\begin{table*}[ht!]
\footnotesize
    \caption{Mined reasoning and corresponding instruction from clinical reports.}
    \label{apptab:mined}
    \setlength{\tabcolsep}{3pt} 
    \centering
    \resizebox{0.95\linewidth}{!}{
    \begin{tabular}{p{1.5cm}|p{3cm}|p{12cm}}
    \toprule
     Task Type &  Instruction & Mined Reasoning \\
    \midrule
    Binary Disease Diagnosis & 
      <Image> Does this chest X-ray show atelectasis? &  The analysis begins by examining the chest X-ray for signs of atelectasis. Key features to evaluate include reduced lung volumes and any distinctive opacities in the lung. Observations revealed low lung volumes and minimal opacities in the left basilar region, suggesting partial segmental collapse indicative of atelectasis. Normally, atelectasis can cause structures like the trachea or diaphragm to shift towards the affected area, but in this case, no significant displacement was noted, supporting the conclusion that the atelectasis is mild. Given these findings, the chest X-ray does indeed show minimal left basilar atelectasis, confirming the diagnosis of atelectasis.
      \\ 
    \midrule
    Single Disease Diagnosis & 
    <image><image> <image>Which finding is in this chest X-ray? Options: A) lung opacity B) support devices C) enlarged cardiomediastinum D) atelectasis  \vspace{1.9cm}
    & To determine the finding in this chest X-ray, we begin by assessing for lung opacity, support devices, enlarged cardiomediastinum, and atelectasis. First, examining lung opacity, we observe mild linear opacity in the bilateral lung bases. This is consistent with bibasilar atelectasis, suggesting that the opacity is related to alveolar collapse rather than fluid accumulation. Moving to support devices, the images reveal multiple overlying leads and linear dense objects in the right upper quadrant, indicative of artifacts from electrocardiographic monitoring and likely cholecystectomy clips. These are not pathological findings but rather support devices. Regarding the mediastinal contour, although there is pulmonary vascular crowding, there is no direct evidence of an enlarged cardiomediastinum. The findings suggest possible mild pulmonary vascular congestion but do not conclusively indicate cardiomegaly or significant enlargement. Finally, the low lung volumes and persistent linear opacities at the lung bases, observed across serial images without significant change, confirm the presence of atelectasis. Given these observations, the primary finding in this chest X-ray is atelectasis. \\
    \midrule
    Multiple Disease Diagnosis & 
    <image>Which findings are in this chest X-ray? Options: A) lung opacity, pneumonia, support devices B) atelectasis, cardiomegaly, support devices C) cardiomegaly, enlarged cardiomediastinum, pneumothorax D) fracture, enlarged cardiomediastinum, atelectasis & First, assess the presence of lung opacity, pneumonia, and support devices. The images reveal a diffuse interstitial pattern and unchanged tracheostomy tube and other support devices. This suggests lung opacities possibly linked to past pneumonia. Next, evaluate the possibility of atelectasis and cardiomegaly along with support devices. There are no visible signs of lung tissue collapse or an enlarged heart. Support devices are noted to be unchanged. Then, consider cardiomegaly, enlarged cardiomediastinum, and pneumothorax. There are no indications of an enlarged heart or mediastinum, nor are there signs of pneumothorax. The status of support devices remains unchanged with previous lung infection sequelae. Finally, check for fracture, an enlarged cardiomediastinum, and atelectasis. No bone fractures or signs of lung collapse are observed. Additionally, there is no evidence of an enlarged mediastinum. In conclusion, the findings that align best with the observations are lung opacity, pneumonia, and support devices. \\
    \midrule
    Anomaly Detection &  
    <image>Which findings are in this chest X-ray? & Upon reviewing the chest X-ray images, increased opacification is observed in the left lower lobe. This pattern is indicative of left basilar atelectasis, where areas of increased density suggest partial or complete lung collapse. Comparing with prior images, the stability of this finding suggests a chronic condition rather than an acute collapse. Additionally, the presence of overlying opacity raises the possibility of superimposed pneumonia, indicating potential infectious consolidation. The images reveal multiple fractures in the left-sided ribs and a displaced fracture in the left clavicle. Given their stability over previous imaging, these injuries are currently in a healing or chronic phase, rather than representing new acute fractures. The consistent appearance across different assessments supports this conclusion. Examining the left basilar region, an overlying opacity is evident at the site of atelectasis. This opacity, in combination with stable atelectasis, aligns with the typical presentation of developing pneumonia, signifying possible infectious consolidation. Inspection of the right apical lung area shows a line suggestive of a small pneumothorax. However, in the context of adjacent fractures, this line is more likely due to displaced rib fragments rather than a genuine pneumothorax. This interpretation requires careful consideration of all imaging details. In conclusion, the findings from the chest X-ray indicate the presence of atelectasis, fracture, and pneumonia. \\
    \midrule
    Temporal Comparison Analysis & 
    <image><image>You are given two images: one reference image and one new image. Please identify the progression of consolidation. Options:A) worsening B) stable C) improving 
    & To assess the progression of consolidation between the two x-ray images, I need to focus on changes in the lung fields, where consolidation manifests as denser areas. Initially, I will observe both images for any variations in opacity across the entirety of the lung regions. Following this, I will pay particular attention to any change in volume or distribution of these opacities, which could indicate progression. Upon carefully comparing the images, I notice that there is an increase in the density and size of the opacities, especially on the left side of the lungs. This indicates that the areas of consolidation have expanded. Given these observations, the conclusion is that there is a worsening of the consolidation, as the increased opacities suggest progression of the condition. \\
    \bottomrule
    \end{tabular}}
\end{table*}

\begin{figure*}[!pht]
    \centering
    \includegraphics[width=0.85\linewidth]{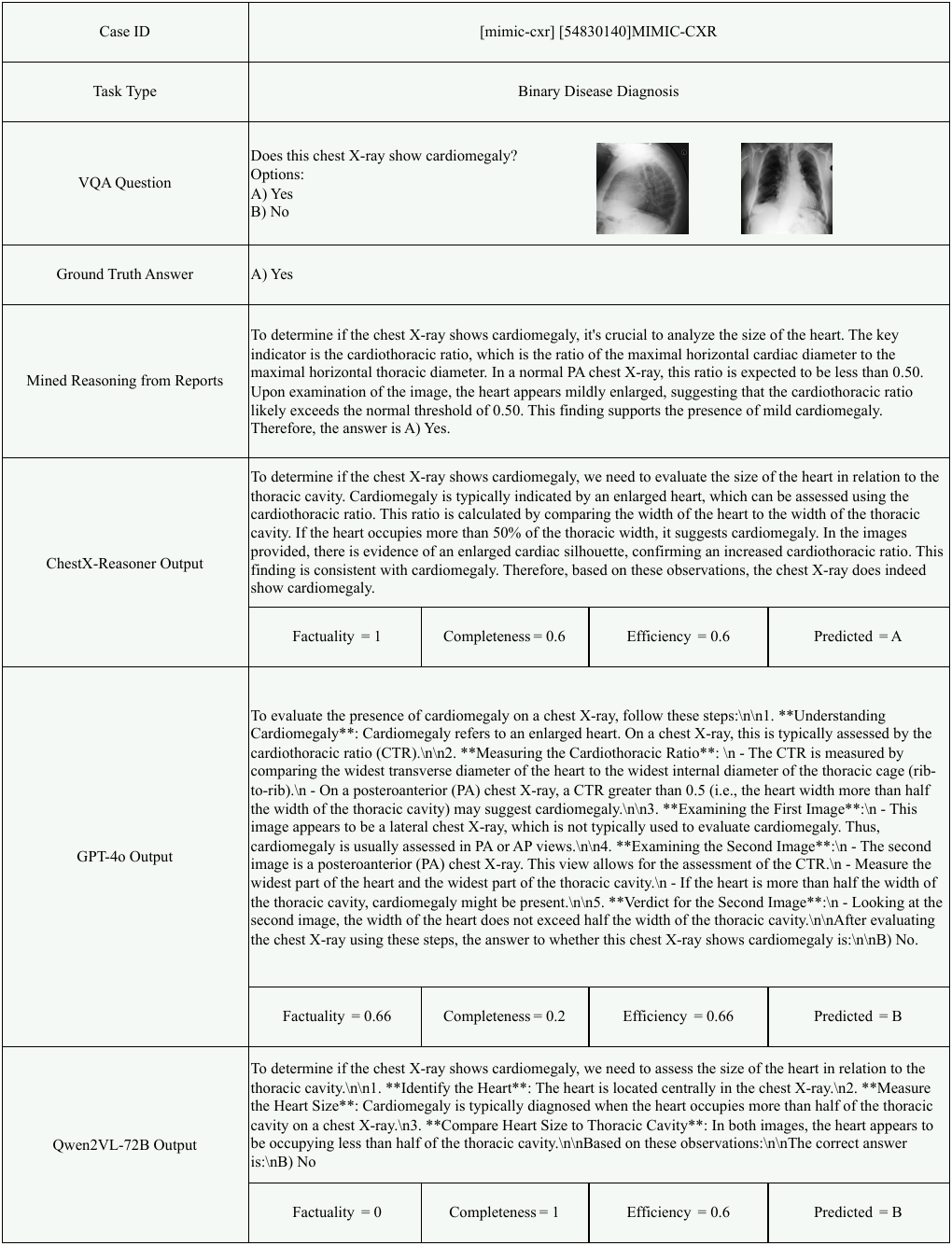}
    \vspace{2pt}
    \caption{Case 1. A case of binary disease diagnosis. \textbackslash n and **xxx** in GPT-4o and Qwen2VL-72B's output represent a line break and bold text for xxx, respectively.
}
    \label{appfig:case1}
    \vspace{-15pt}
\end{figure*}

\begin{figure*}[!pht]
    \centering
    \includegraphics[width=0.85\linewidth]{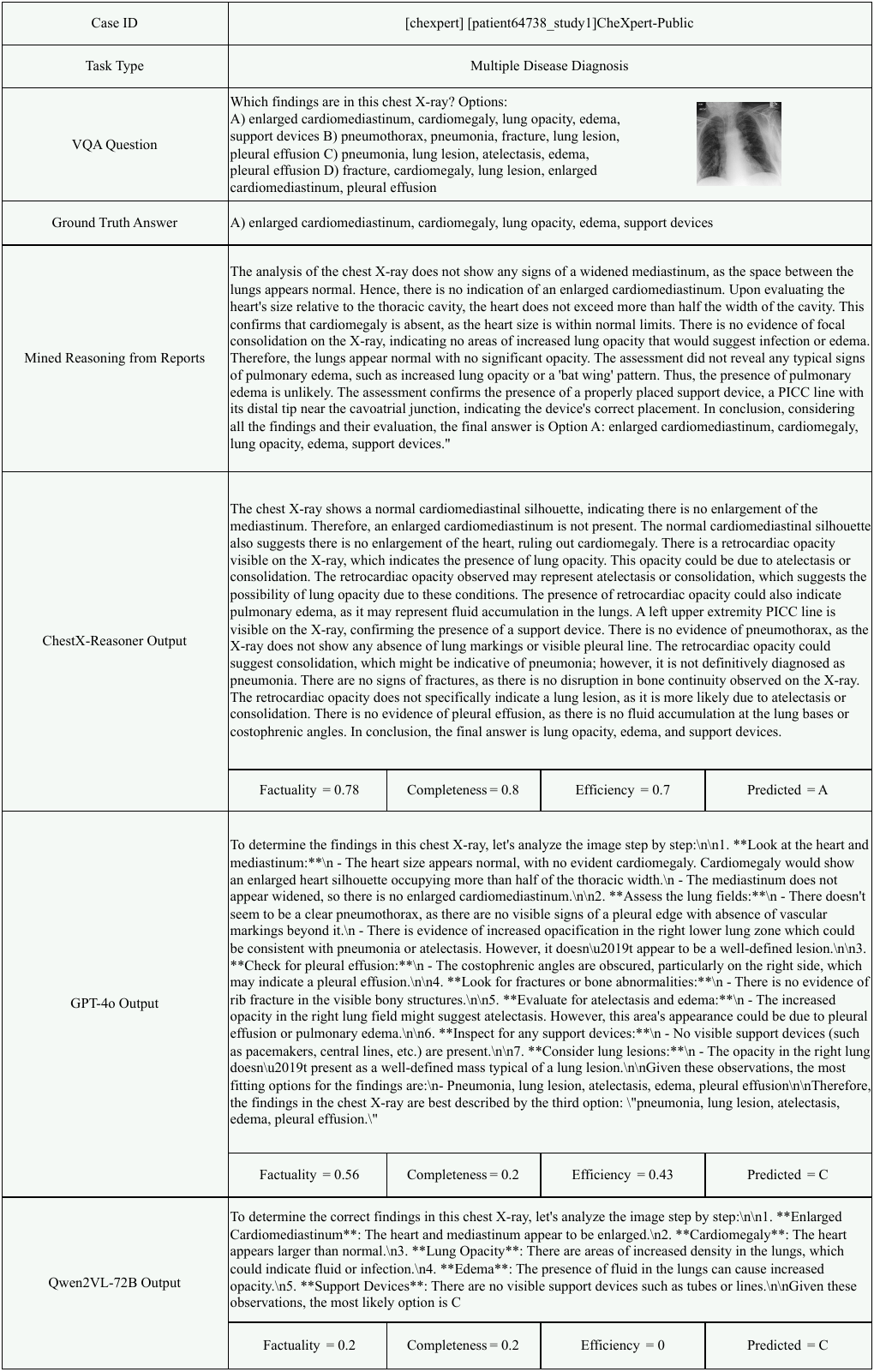}
    \vspace{2pt}
    \caption{Case 2. A case of multiple disease diagnosis. \textbackslash n and **xxx** in GPT-4o and Qwen2VL-72B's output represent a line break and bold text for xxx, respectively.}
    \label{appfig:case2}
    \vspace{-15pt}
\end{figure*}

\begin{figure*}[!pht]
    \centering
    \includegraphics[width=0.85\linewidth]{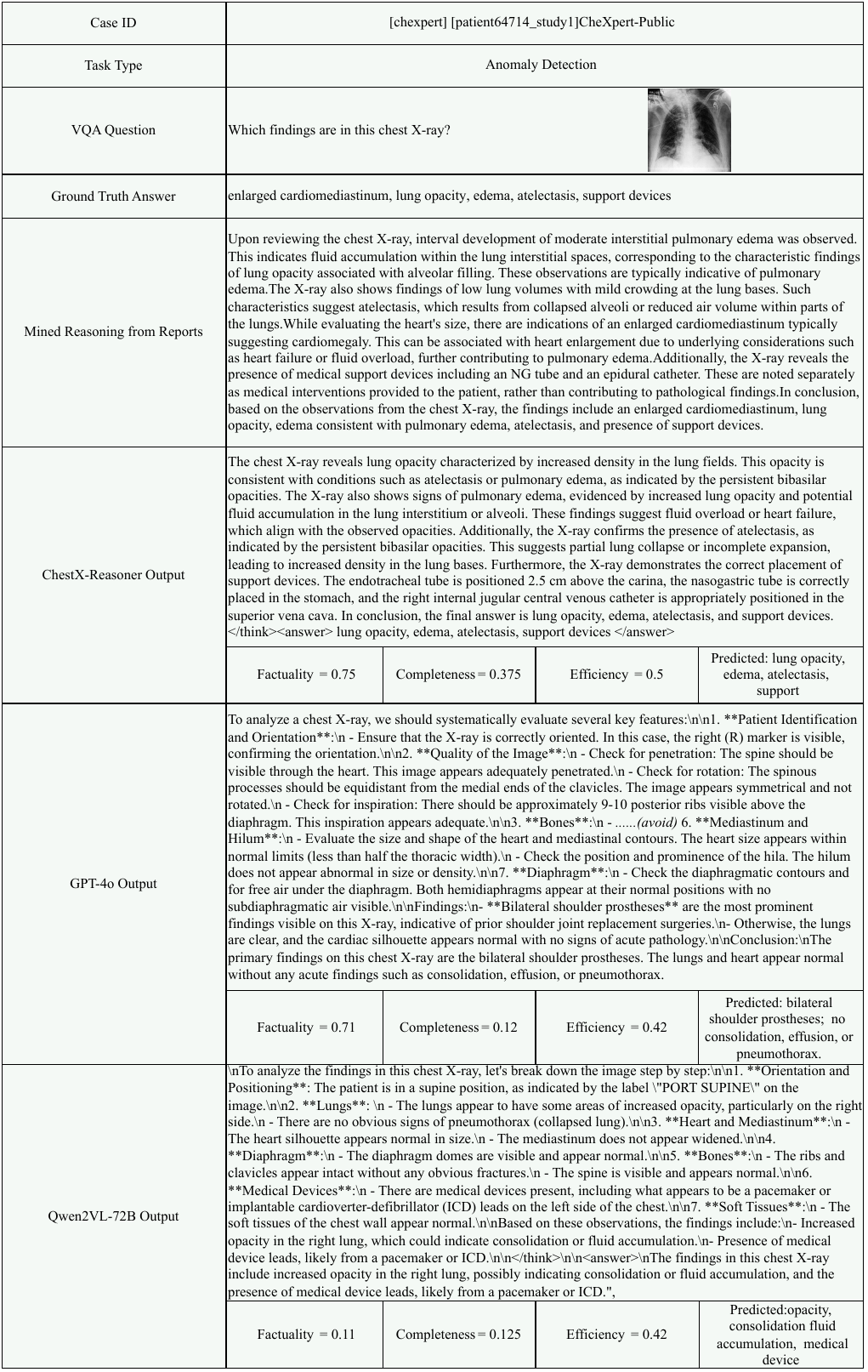}
    \vspace{2pt}
    \caption{Case 3. A case of anomaly detection. \textbackslash n and **xxx** in GPT-4o and Qwen2VL-72B's output represent a line break and bold text for xxx, respectively.}
    \label{appfig:case3}
    \vspace{-15pt}
\end{figure*}

\begin{figure*}[!pht]
    \centering
    \includegraphics[width=0.85\linewidth]{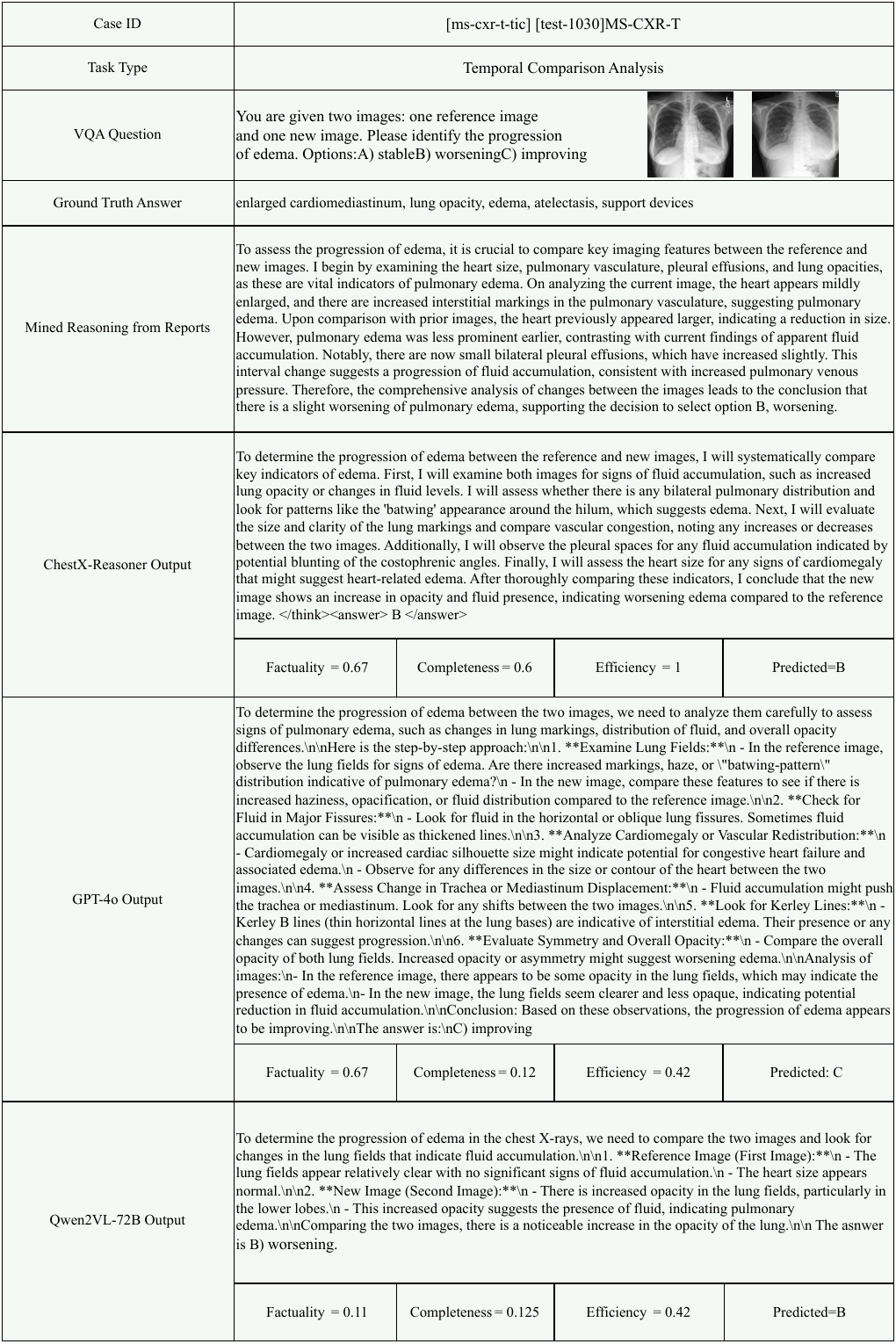}
    \vspace{2pt}
    \caption{Case 4. A case of temporal comparison analysis. \textbackslash n and **xxx** in GPT-4o and Qwen2VL-72B's output represent a line break and bold text for xxx, respectively.}
    \label{appfig:case4}
    \vspace{-15pt}
\end{figure*}

\subsection{Training Details}\label{app:training}
We utilize Qwen2VL-7B~\cite{qwen2vl} as our base model due to its strong reasoning performance in the domains of mathematics and programming.
As shown in Supplementary Table~\ref{apptab:param1}, the training parameters for the first supervised fine-tuning stage are configured as follows: the optimizer is AdamW~\cite{adamw}, with $\beta_1$ set to 0.9 and $\beta_2$ to 0.95. We adopt a cosine learning rate schedule with an initial learning rate of 2e-6, a batch size of 512 samples, a weight decay of 0.0, and a gradient clipping threshold of 1.
The experiments are conducted on 8×Tesla A100 GPUs using DeepSpeed-ZeRO2~\cite{deepspeed}. In this stage, the model is trained for 3500 steps, taking approximately 2 days to complete.
For the second stage, reinforcement learning with process reward, the training parameters are provided in Supplementary Table~\ref{apptab:param2}.
The optimizer remains the same as in the first stage, with a learning rate of 5e-7. Both the global batch size and mini updating batch size are set to 256. The coefficient of KL loss is 0.0001, and the group number ($G$) is 8. The reward function combines outcome format reward, outcome correctness reward, and process reward, while entropy loss is 0.0001.
The experiments are conducted on 8×Tesla A100 GPUs using PyTorch-FSDP~\cite{fsdp} and VeRL engine~\cite{verl}. In this stage, the model is trained for 400 steps, taking approximately 3.5 days to complete.

\subsection{Instruction Design}\label{app:instruction}
For the model to operate in both reasoning mode and direct response mode, we adopt two instruction formats for each sample, each paired with its corresponding expected output. 

\begin{itemize}
\vspace{-8pt}
    \item \textbf{Instruction without Chain of Thoughts Prompt:} 
    \begin{quote}
   (Input Instruction)~System: You are a helpful AI assistant. \\
    (Input Instruction)~User: <image> ... <image>Question. Please enclose the answer within <answer></answer> \\
(Expected Output Answer)~Assistant: <answer> option etc. B or open-ended answer </answer>. \\
\vspace{-10pt}
    \end{quote}
    \item \textbf{Instruction with Chain of Thoughts Prompt:} 
    \begin{quote}
    (Input Instruction)~System: You are a helpful AI assistant. \\
(Input Instruction)~User: <image> ... <image>Question. Please think step by step, and enclose the answer within <answer></answer> and the reasoning processes within <think></think>. \\
(Expected Output Answer)~<think> First, assess the presence of consolidation at bilateral lung, ..., Finally, check for Pleural nodule on left side, ..., In conclusion, the findings that align best with the observations are B) Sutures in
upper lobe of left lung, Cicatrix on left ribs </think><answer> option etc. B or open-ended answer </answer>. \\
    \end{quote}
\vspace{-8pt}
\end{itemize}

\subsection{Mined Reasoning Visualization}\label{app:mined}
As shown in Supplementary Table~\ref{apptab:mined}, we show the task type, input instruction, and mined reasoning processes of cases in our benchmark RadRBench-CXR.

\subsection{Output Reasoning Visualization}\label{app:output}
To better demonstrate the effectiveness of our method and the comparison of reasoning ability of the model, we present four cases of reasoning outputs from our ChestX-Reasoner, GPT-4o, and Qwen2VL-72B, alongside the corresponding questions, answers, and extracted reasoning processes, as shown in Supplementary Fig.~\ref{appfig:case1},\ref{appfig:case2},\ref{appfig:case3}, and~\ref{appfig:case4}.

\clearpage

\end{document}